\documentclass[conference]{IEEEtran}

\usepackage[utf8]{inputenc}
\usepackage[T1]{fontenc}
\usepackage[english]{babel}
\usepackage{subscript}
\usepackage[table]{xcolor}

\usepackage{booktabs} 
\usepackage{array}

\usepackage{graphicx, float, subfigure, blindtext}

\newcommand\IEEEhyperrefsetup{
bookmarks=true,bookmarksnumbered=true,%
colorlinks=true,linkcolor={black},citecolor={black},urlcolor={black}%
}

\usepackage[\IEEEhyperrefsetup, pdftex]{hyperref}

\usepackage{mathtools}

\usepackage[nolist,nohyperlinks]{acronym}
\usepackage{cleveref}
\graphicspath{{images/}}

\usepackage{titlesec}
\titlespacing*{\section}{0pt}{*1}{*1}
\titlespacing*{\subsection}{0pt}{*1}{*1}
\renewcommand{\thesubsubsection}{\arabic{subsubsection}}
\titleformat{\subsubsection}[runin]{\itshape}{\thesubsubsection)}{1em}{}[:]
\titlespacing*{\subsubsection}{\parindent}{0pt}{*1}

\author{\textbf{\href{https://www.jua.ai/}{Jua.ai | AI for weather-dependent energy trading}}\\
\IEEEauthorblockA{%
Roberto Molinaro\textsuperscript{*}, Jordan Dane Daubinet\textsuperscript{*},\\
Alexander Jakob Dautel,  Andreas Schlueter, Alex Grigoryev, Nikoo Ekhtiari, Bas Steunebrink, \\
Kevin Thiart, Roan John Song, Henry Martin, Leonie Wagner, Andrea Giussani,  \\
Marvin Vincent Gabler (all affiliated with Jua.ai\textsuperscript{1})\\
\vspace{0.5em}
\textsuperscript{*}\textit{These authors contributed equally to this work.}}%
}

\title{EPT-1.5 Technical Report}
\begin{document}
\maketitle
\pagestyle{plain}
\thispagestyle{plain} 
\footnotetext[1]{\url{https://www.jua.ai/}}
\begin{abstract}
We announce the release of \mbox{EPT-1.5}, the latest iteration in our Earth Physics Transformer (EPT) family of foundation AI earth system models. \mbox{EPT-1.5} demonstrates substantial improvements over its predecessor, \mbox{EPT-1}. Built specifically for the European energy industry, \mbox{EPT-1.5} shows remarkable performance in predicting energy-relevant variables, particularly 10m \& 100m wind speed and solar radiation. Especially in wind prediction, it outperforms existing AI weather models like GraphCast, FuXi, and Pangu-Weather, as well as the leading numerical weather model, IFS HRES by the European Centre for Medium-Range Weather Forecasts (ECMWF), setting a new state of the art. \\
\end{abstract}
\section{Introduction}
\label{section:intro}

Weather forecasting has dramatically evolved from its early reliance on empirical observations to the sophisticated numerical models that form the backbone of modern meteorology. The foundation of numerical weather prediction was established by Vilhelm Bjerknes in the early 20th century. Bjerknes proposed that atmospheric processes could be represented by a set of mathematical equations—specifically, the fundamental laws of physics governing fluid dynamics and thermodynamics. By numerically solving these equations, it became possible to predict future states of the atmosphere based on current observational data.

Today, numerical weather prediction models simulate the atmosphere by dividing it into a three-dimensional grid and calculating the changes in atmospheric variables like temperature, pressure, wind speed, and humidity at each grid point over time. These models require immense computational power and are continually refined to include more complex processes, such as cloud formation, radiation, and interactions between the atmosphere and the Earth's surface. Advanced data assimilation techniques incorporate real-time observational data from satellites, weather stations, and radar to enhance the accuracy of these models. Alongside, artificial intelligence and machine learning have been increasingly employed to optimize model parameters and improve predictive capabilities, marking a new era in weather forecasting.

Building upon these advancements, foundation AI Earth system models have emerged that directly map raw observational data to predictions using end-to-end techniques. These models leverage deep learning to process vast datasets from satellites, radars, and sensors, capturing complex interactions among the atmosphere, oceans, land surfaces, and biosphere. By bypassing some traditional numerical methods, machine learning models aim at providing faster and more accurate forecasts. While challenges remain in data accessibility and model interpretability, these end-to-end approaches represent a significant step toward a more holistic and efficient understanding of Earth's systems.

The current report is meant to elucidate the development of Jua's current \mbox{EPT-1.5} model. It represents a minor update to the previous version, \mbox{EPT-1}, including only small changes in architecture and training strategy, while maintaining comparable  size. Nevertheless, it shows substantial performance improvements compared to its predecessor. Like \mbox{EPT-1}, \mbox{EPT-1.5} leverages a proprietary variant of the transformer architecture\cite{vaswani2023attentionneed}, widely recognized for its performance in both vision and language models. The key strength of this architecture lies in its ability to scale effectively with increasing model size and data.

\section{AI Weather Forecasting at Jua}
\label{section:jua-hist}

Since its inception, Jua is dedicated to building a physically consistent world model to simulate our Earth and predict the future. Our journey includes several milestones:

\subsection{World's First Operational Global AI Weather Forecasting Model}

On March 1, 2023, Jua launched Vilhelm \cite{gabler2023vilhelmAIweather}, the world's first operational global AI weather forecasting model. This marked a special moment in meteorology, demonstrating the potential of AI to revolutionize weather prediction. Jua Vilhelm also set a new standard by being the first global AI weather model capable of performing predictions with an hourly temporal resolution, in contrast to 6-hourly predictions by other research AI models.

\subsection[First Global 1~x~1~km Resolution Precipitation Forecast]
{First Global 1~$\times$~1~km Resolution Precipitation Forecast}
One month later, on April 1, 2023, Jua introduced the world's first \textit{operational} global precipitation forecast with a 1~$\times$~1~km resolution \cite{gabler2023highresAIprecip}. This ultra-high-resolution model provides unparalleled detail in precipitation predictions. To the best of our knowledge, this model's capability has not been replicated by any other lab yet. 
\section{EPT-1.5: Model Architecture and Innovations}
\label{section:arch}

\mbox{EPT-1.5} represents the latest evolution in Jua's AI weather forecasting capabilities. It is a foundation earth systems model based on Jua's EPT-1 architecture.

\subsection{Key Features and Innovations}

\begin{enumerate}
    \item \textbf{Foundation Earth Systems Model}: Like its predecessor \mbox{EPT-1}, \mbox{EPT-1.5} is designed to solve partial differential equations (PDEs), the fundamental building blocks of atmospheric and oceanic dynamics. It is an AI foundation model and trained on a diverse dataset with varying spatial and temporal resolution, as well as varying input and output weather parameters.
    
    \item \textbf{Any Lead Time Forecasting}: Unlike most current AI weather models that can only provide forecasts every six hours, \mbox{EPT-1.5} offers predictions for any future lead time. This allows for accurately simulating the exact time of a specific event. While existing numerical or AI weather models cannot simulate global weather in a high temporal resolution due to computational limits, \mbox{EPT-1.5} can generate native minute-by-minute forecasts extending days or weeks into the future in an operational setting.
    
    \item \textbf{Large-Scale Model and Dataset}: With a parameter count in the billions, \mbox{EPT-1.5} has the capacity to capture intricate weather patterns and relationships across global scales.
    The model is trained on 5 petabytes of weather data, ensuring comprehensive coverage of diverse atmospheric conditions and phenomena. Building and maintaining the data infrastructure was one of the biggest challenges in building the EPT series. While we trained smaller versions of the model for scaling experiments, these models were not operationalized.
    
    \item \textbf{Diverse Finetuning Techniques}: After pretraining our foundation model, we employ various finetuning methods to enhance model stability and performance, ensuring reliable predictions across various weather scenarios. These techniques include improvements of the long term performance for weeks out and improvements on model robustness to ensure reliable daily operational performance.

    \item \textbf{Probabilistic Forecasting}: We implement probabilistic capabilities for our models in order to predict multiple possible future scenarios and increase long term performance. However, all benchmarks in this report are purely deterministic for comparability with other AI models.
    
\end{enumerate}

\subsection{Operational Specifications}

\begin{itemize}
    \item \textbf{Spatial Resolution}: \mbox{EPT-1.5} operates at a spatial resolution of 0.083 degrees or roughly 9 x 9 km at the equator, providing highly detailed forecasts.
    \item \textbf{Temporal Resolution}: \mbox{EPT-1.5} currently runs at a temporal resolution of 1 hour up to 20 days into the future.
    \item \textbf{Forecast Frequency}: The model runs four times daily at 00:00, 06:00, 12:00, and 18:00 UTC. At each of these times, both an early version and a standard version are executed. The early version provides predictions up to 3 hours ahead of other models like the ECMWF's IFS, while the standard version incorporates more input data but operates with a 6-hour delay, which is currently common in the weather industry.
\end{itemize}
\section{EPT Model Variants}
\label{section:method}

The predominant paradigm of machine learning in the context of weather forecasting involves predicting the future by mapping an initial state to a future state. Longer term forecast are then obtained by autoregressively feeding the model with its own predictions. 

This process can be more rigorously formalized by approximating the mapping:
\[
\Tilde{X}^{t+\Delta t} = f(X_t, X_{t-1}, \dots),
\]
which delineates the relationship between the current and previous states of the weather and the future state through a machine learning model. In this formulation, the model is denoted by $f_\theta$, where:
\[
X^{t+\Delta t} \approx \Tilde{X}^{t+\Delta t} = f_\theta(X_t, X_{t-1}, \dots).
\]

Due to the complexity of this task, we develop distinct variants of \mbox{EPT-1.5} tailored to address the nuances of each specific subtask involved.

\vspace{0.5cm}
\textbf{EPT\textsubscript{Base}}: EPT\textsubscript{Base} refers to the pretrained foundation model. 

\vspace{0.5cm}

\textbf{EPT-1.5\textsubscript{$\alpha$}}:
EPT-1.5\textsubscript{$\alpha$} is based on EPT\textsubscript{Base}  and represents a model specifically finetuned for robust long term predictions.

\vspace{0.5cm}

\textbf{EPT-1.5\textsubscript{Op}}: EPT-1.5\textsubscript{Op} denotes the operational model which is finally deployed in production.
\vspace{0.5cm}
\section{Benchmarks Details}
\label{section:benchmarking}
In this section, we outline the evaluation process for the performance of the EPT models. The pipeline is based on the well-established WeatherBench \cite{wb}.

\subsection{Ground Truth Datasets}
A fundamental aspect of evaluating model performance is the definition of a ground truth dataset. In the following, we describe three accepted datasets in the weather community.

\textbf{ERA5.}
ERA5\cite{era5dataset} is a state-of-the-art global atmospheric reanalysis dataset produced by the European Centre for Medium-Range Weather Forecasts (ECMWF). It is a gridded dataset and provides hourly estimates of a wide array of atmospheric, land, and oceanic variables from 1979 to the present, with a horizontal resolution of approximately 31~x~31 km at the equator (0.25 degrees). ERA5 assimilates vast amounts of historical observational data using ECMWF's Integrated Forecasting System (IFS), ensuring a consistent and detailed representation of the Earth's climate and weather patterns over the past decades.

\textbf{IFS HRES IC.}
The ECMWF High-Resolution Forecast (HRES) initial conditions (IC) are computed from the most recent observational data, using a smaller data assimiliation window compared to ERA5. This dataset is commonly known as HRES Analysis and represents a gridded dataset. The high resolution initial conditions serve as the starting point for ECMWF's deterministic weather forecasts, which operate at a horizontal resolution of about 9~x~9 km at the equator (0.1 degrees).

\begin{figure}
  \centering
  \includegraphics[width=\columnwidth]{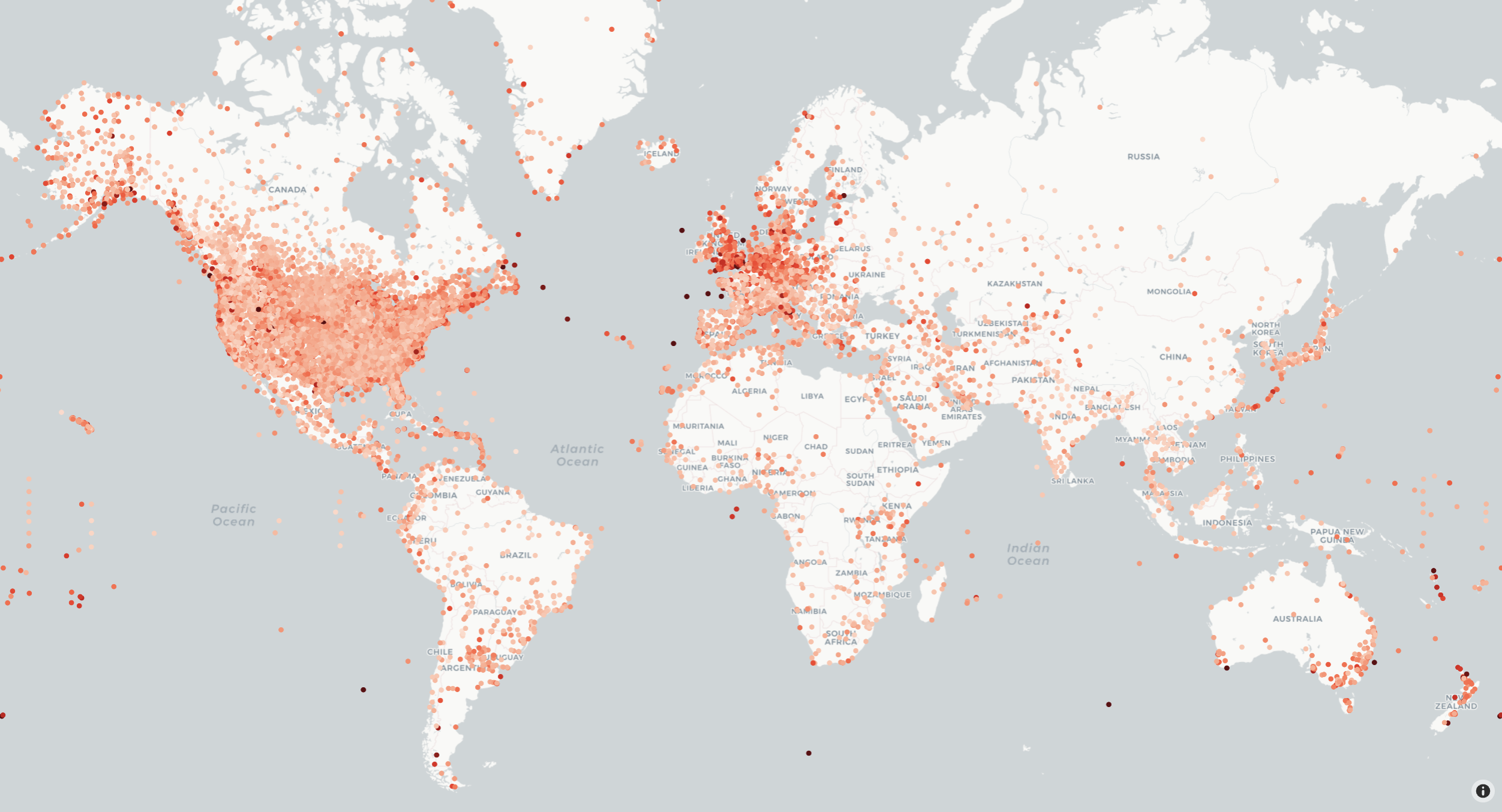}
  \caption{Spatial distribution of the weather stations for benchmarking}
  \label{fig:point_stations}
\end{figure}

\textbf{Weather Stations.}
SYNOP (Surface Synoptic Observations) and METAR (Meteorological Aerodrome Reports) are internationally standardized codes used for reporting weather observations from land-based meteorological stations and airports, respectively. SYNOP data are collected multiple times per day and include a wide range of atmospheric parameters, providing essential information for synoptic-scale weather analysis and forecasting. METAR reports are also issued multiple times a day, offering real-time weather details critical for aviation operations, including all common surface weather variables. As these stations have their sensors mounted in heights between 2m and 10m above ground, they do not collect wind measurements in 100m (see Figure \ref{fig:point_stations} for a visual representation of the spatial distribution of the weather stations). 

Both SYNOP and METAR have their main focus on the EU and the US. In addition to these datasets, we purchased proprietary weather station data to increase the number of weather stations to benchmark against from 20,000 to over 100,000. By combining these observation sources into a single point dataset, we ensure a dense global coverage.

\subsection{Benchmarking Approach}
We continue by describing two different types of models benchmarking.

\begin{enumerate}
    \item \textbf{Grid Benchmarks}: This approach compares the gridded model outputs with a gridded ground truth dataset, specifically ERA5 and IFS HRES IC. While these datasets are only an approximation of the ground truth and generated using numerical simulation, they provide data for most points, variables and vertical height layers globally. 
    \item \textbf{Point Benchmarks}: This approach compares the model output at a specific point with an actual observation from a weather station. While this approach is regarded as the gold standard for accuracy, many AI weather models do not publish benchmarks against point observations, making direct comparison difficult.
\end{enumerate}

\subsection{Methodology}
Next, we describe our benchmarking methodology more precisely.

\textbf{Initial Conditions.}
To ensure comparability, EPT-1.5 is initialized with the same data other models used. This approach ensures a standardized baseline for model comparison, although it's important to note that ERA5 data is not available in real-time operational settings. This benchmarking method, while academically rigorous, differs from actual operational conditions. To indicate the operational performance of EPT-1.5, we also benchmark against weather stations and initialize EPT-1.5 with IFS initial conditions, which are available for daily operations.

\textbf{Reference Forecast.}
As reference, we choose the IFS HRES forecast produced by the European Centre for Medium-Range Weather Forecasts (ECMWF) \cite{ecmwf}, benchmarked against its initial conditions and weather stations. IFS HRES is currently considered to be the world's most accurate numerical weather model. We use ground truth data from 2023 to benchmark EPT-1.5 and HRES.

\textbf{Correct Initial Error.}
Some earlier papers benchmarked HRES forecasts against ERA5 as the ground truth, resulting in an initial RMSE greater than 0 at timestep \( t = 0 \), since HRES is not initialized with ERA5 data. This method skews the performance results in favor of the AI model and against HRES. We chose not to use this questionable approach and instead conducted a clean evaluation, including point benchmarking against weather stations, the gold standard in benchmarking for weather forecasts.

By using consistent initial conditions and corresponding ground truth datasets for each model, we ensure that our benchmarking results are scientifically and practically valid.

\textbf{Skill Score.}
Performance is evaluated based on the skill score (SS) defined as

    $$
    \text{SS} = 1 - \frac{\text{RMSE}_{\text{model}}}{\text{RMSE}_{\text{reference}}},
    $$
    with RMSE denoting the Root Mean Square Error:
    $$
    \text{RMSE} = \sqrt{{\sum_{i=1}^{n} w_i \left( \tilde{X}_i - X_i \right)^2}}.
    $$
 Here, \( \tilde{X}_i \) represents the predicted value, \( X_i \) is the weather state corresponding to the ground truth dataset, while the reference forecast is IFS HRES as outlined above. In simple words, skill score is a comparison of how much a model improves over HRES. Thus, a lower RMSE and a higher skill score indicate a better weather forecast.

\section{Results}
\begin{figure}[!t]
  \centering
  \includegraphics[width=\columnwidth]{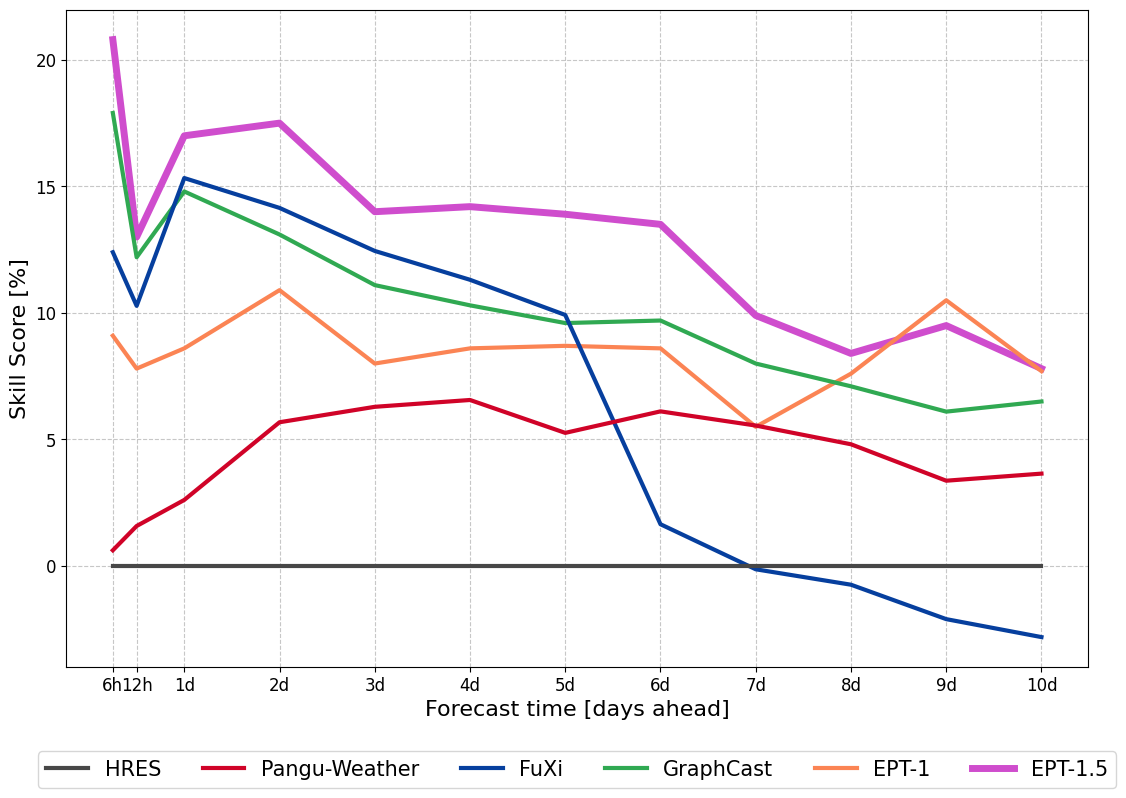}
  \caption{Comparison of 10m wind speed skill score of \mbox{EPT-1.5\textsubscript{$\alpha$}} (pink), \mbox{EPT-1} (orange), GraphCast (green), Pangu (red) and FuXi (dark blue) vs. HRES (black) over Europe}
  \label{fig:10m_wind_speed}
\end{figure}

\begin{figure}[!t]
  \centering
  \includegraphics[width=\columnwidth]{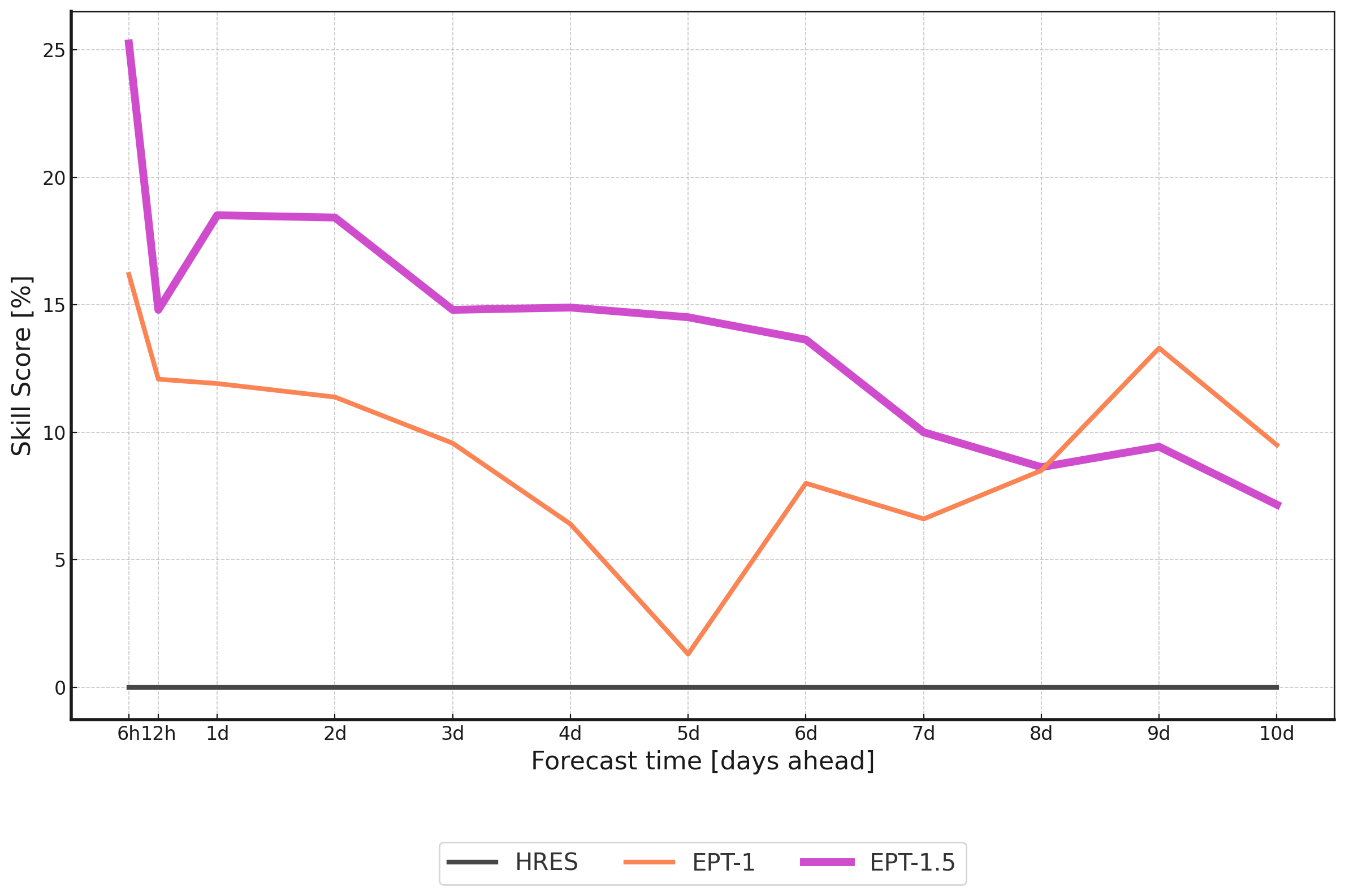}
  \caption{Comparison of 100m wind speed skill score of \mbox{EPT-1.5\textsubscript{$\alpha$}} (pink), \mbox{EPT-1} (orange) vs. HRES (black) over Europe}
  \label{fig:100m_wind_speed}
\end{figure}

\begin{figure}[!t]
  \centering
  \includegraphics[width=\columnwidth]{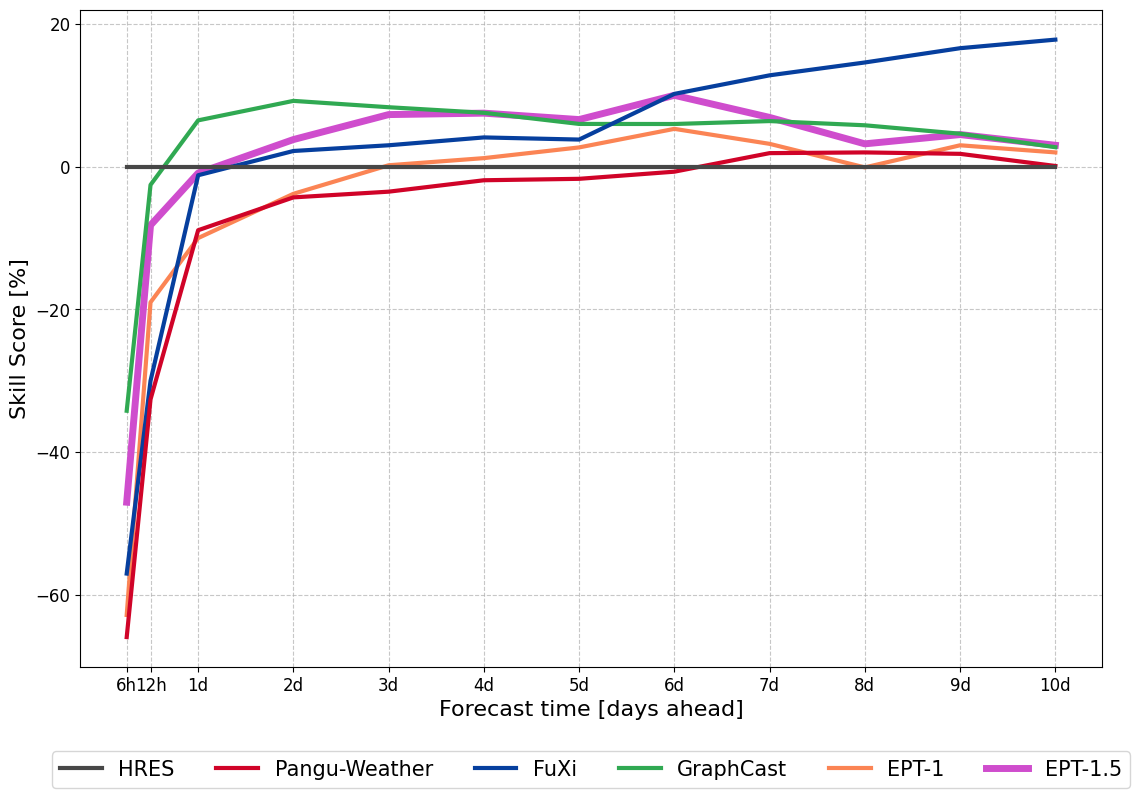}
  \caption{Comparison of 2m temperature skill score of \mbox{EPT-1.5\textsubscript{$\alpha$}} (pink), \mbox{EPT-1} (orange), GraphCast (green), Pangu (red) and FuXi (dark blue) vs. HRES (black)  over Europe}
  \label{fig:2m_temperature}
\end{figure}

\begin{figure}[!t]
  \centering
  \includegraphics[width=\columnwidth]{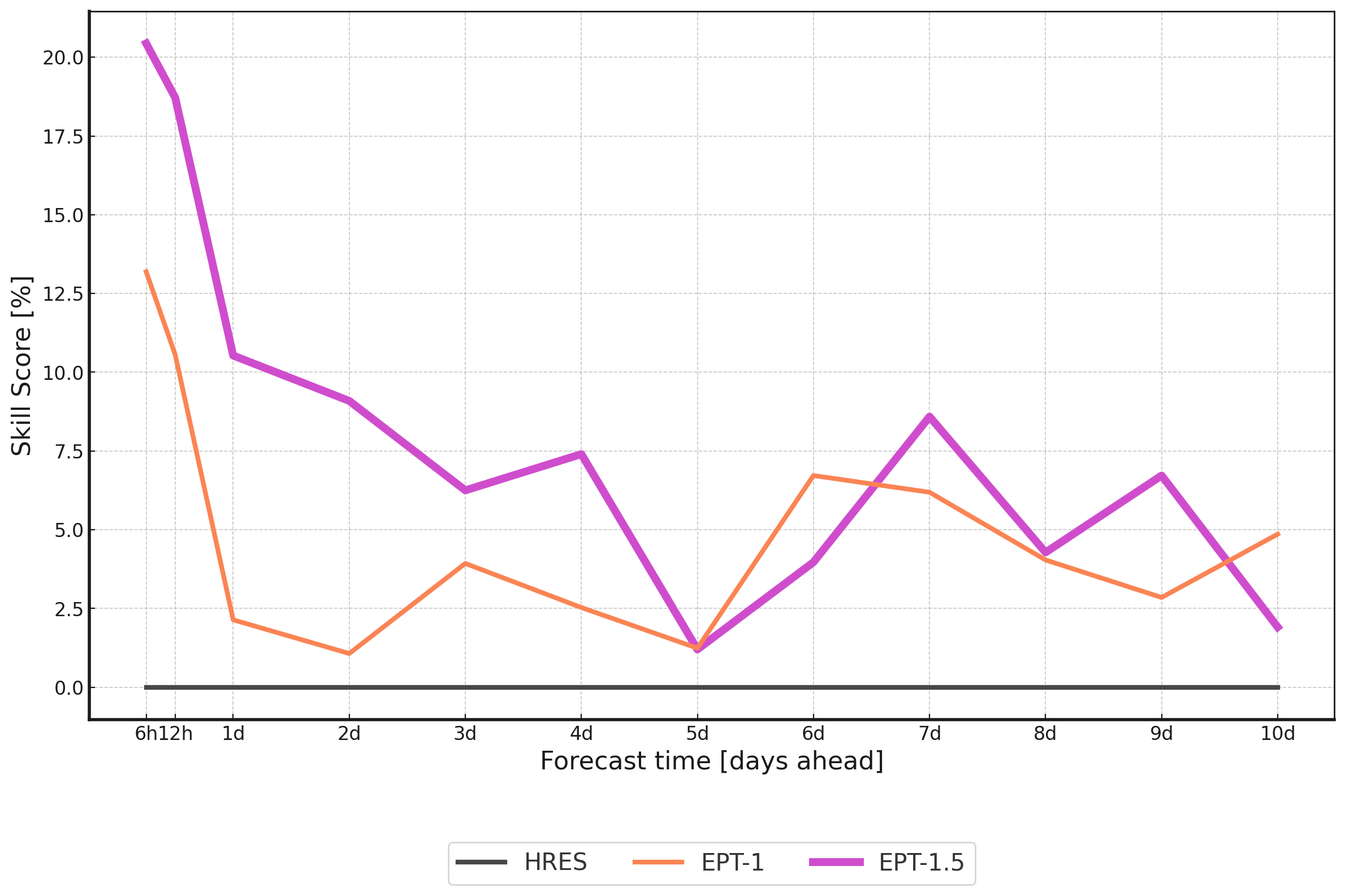}
  \caption{Comparison of 6-hourly surface (downward) solar radiation skill score of \mbox{EPT-1.5\textsubscript{$\alpha$}} (pink), \mbox{EPT-1} (orange) vs. HRES (black) over Europe}
  \label{fig:6h_solar}
\end{figure}

\label{section:results}
In this section, we show and compare the performance of the \mbox{EPT-1.5} family with its predecessor, \mbox{EPT-1}, as well as with three major competing models: GraphCast \cite{lam2023graphcastlearningskillfulmediumrange}, FuXi \cite{chen2023fuxicascademachinelearning}, and Pangu-Weather \cite{bi2022panguweather3dhighresolutionmodel}. The data for the models are retrieved from Weatherbench \cite{wb}, where they are the best performing models.

\begin{figure*}[!ht]
\centering
\begin{tabular}{cc}
    \includegraphics[width=0.45\textwidth]{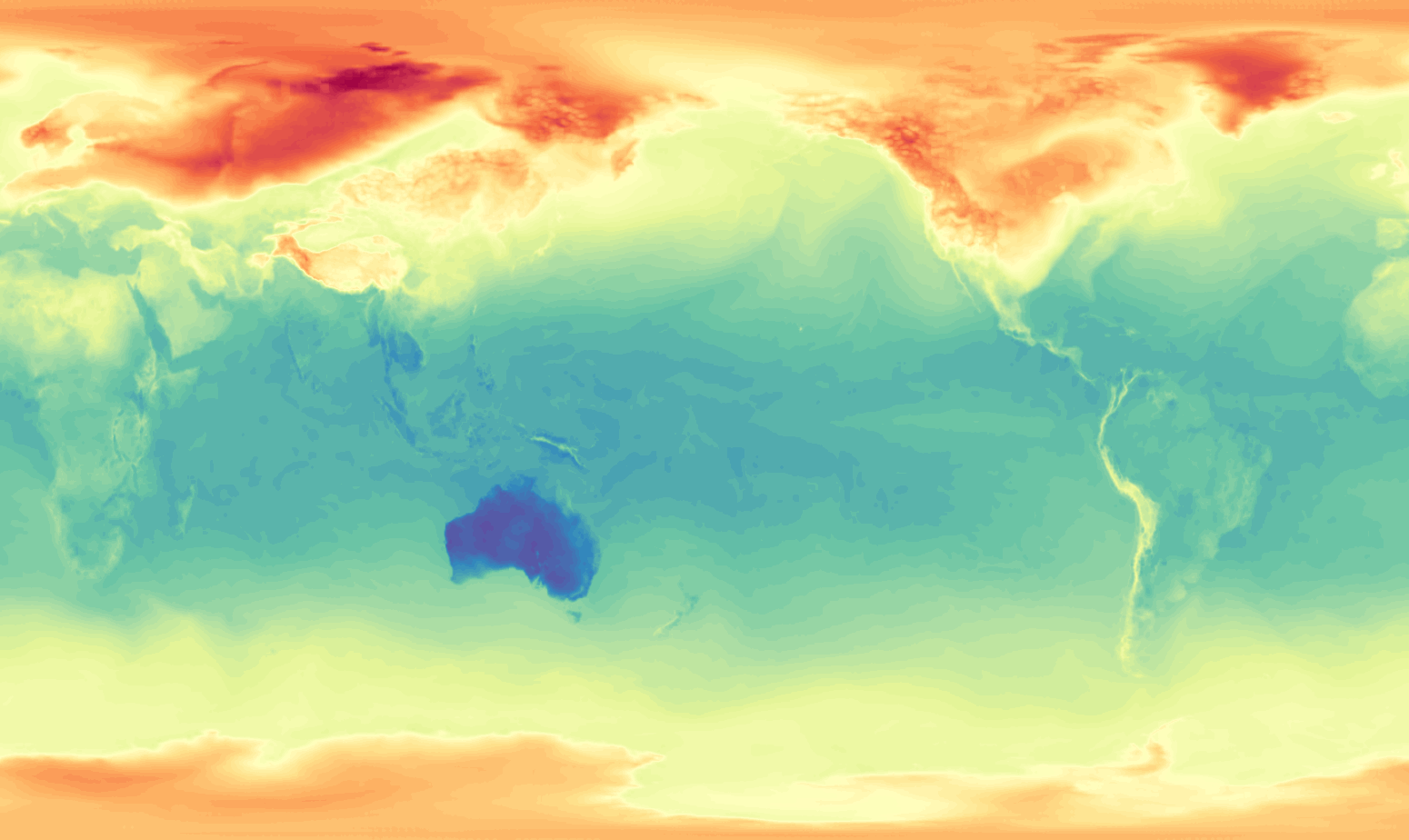} &
    \includegraphics[width=0.45\textwidth]{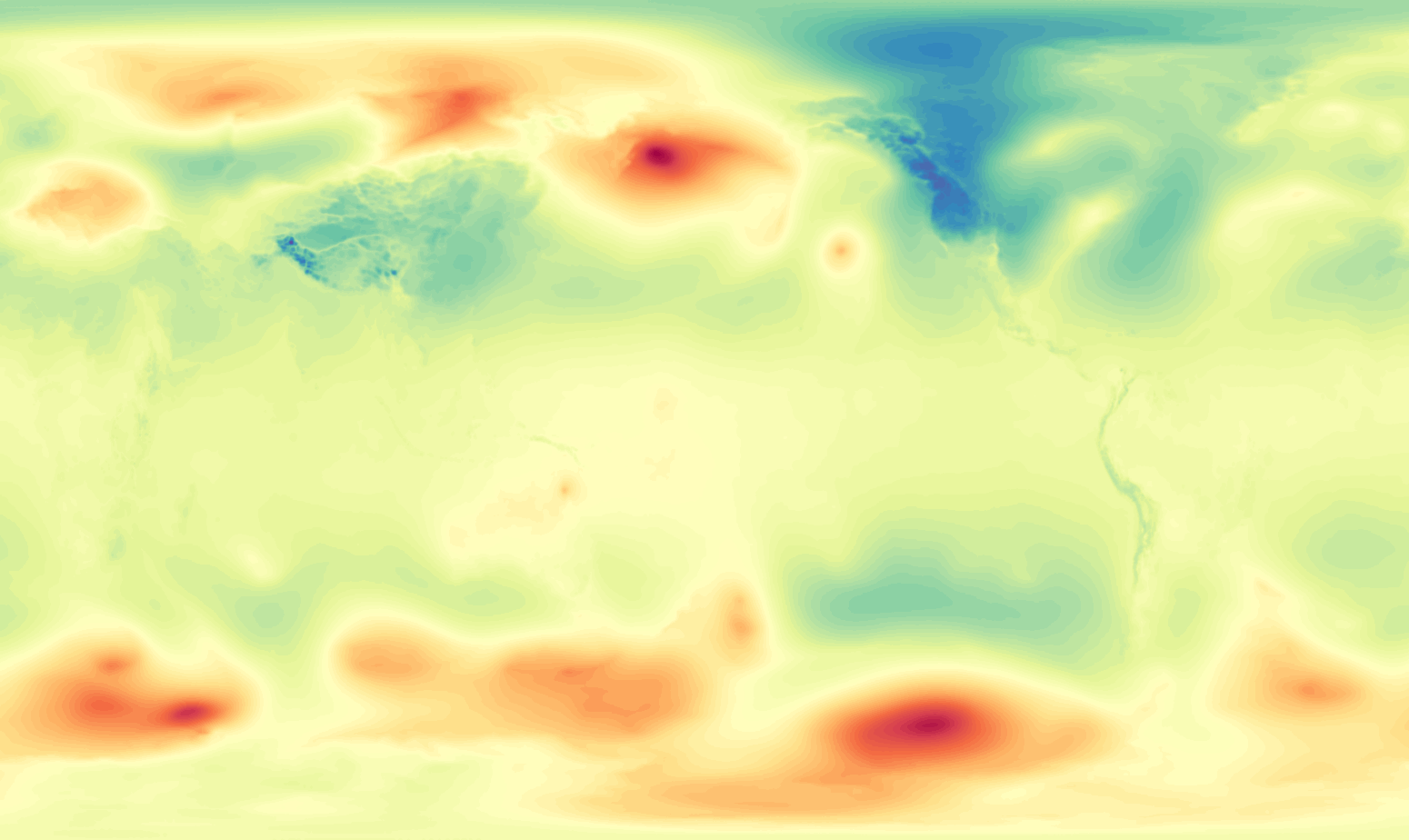} \\
    \small (a) 2m temperature prediction & \small (b) Mean sea level pressure prediction \\[6pt]
    
    \includegraphics[width=0.45\textwidth]{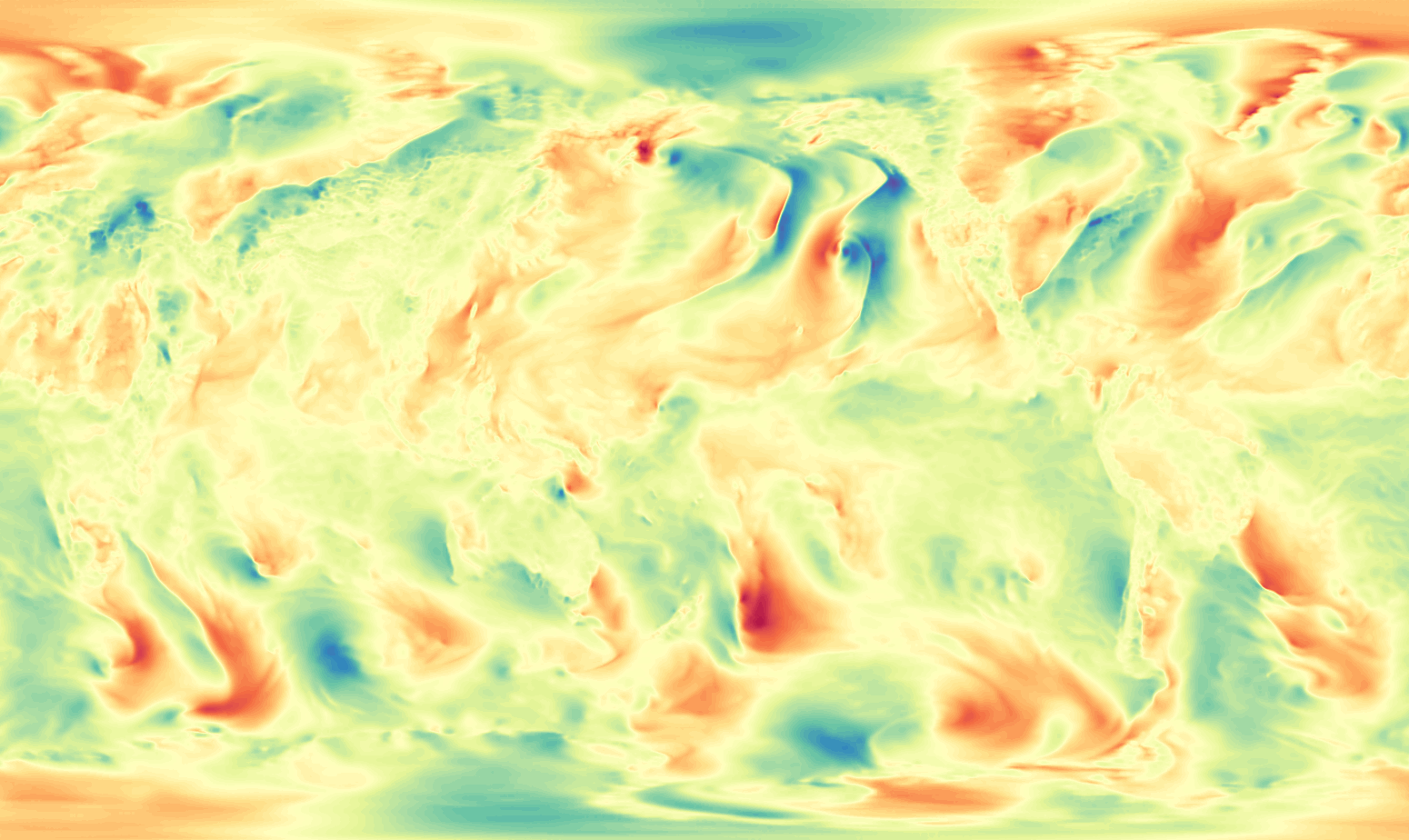} &
    \includegraphics[width=0.45\textwidth]{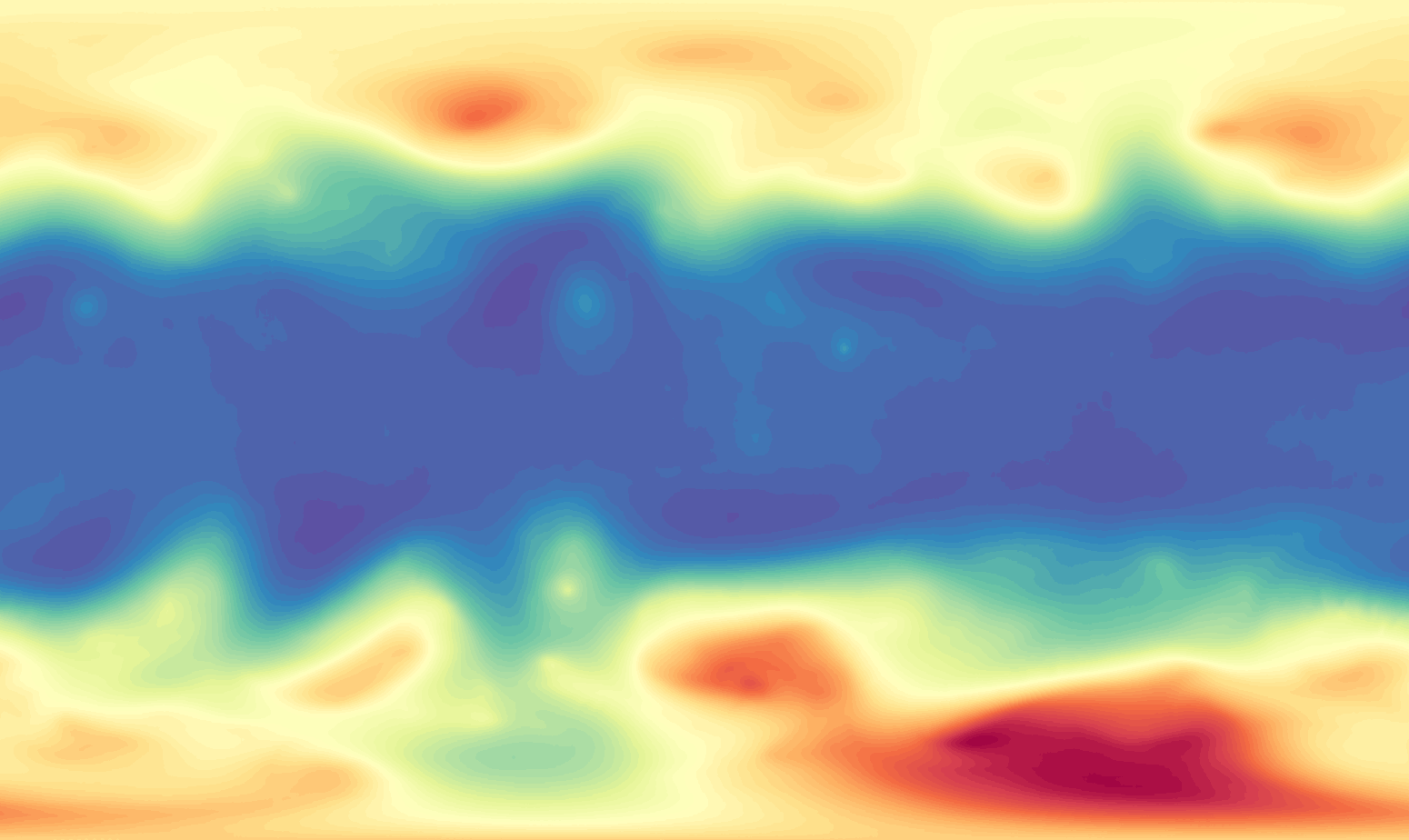} \\
    \small (c) 100m wind speed eastward prediction & \small (d) Geopotential prediction at 500 hPa \\[6pt]
    
    \includegraphics[width=0.45\textwidth]{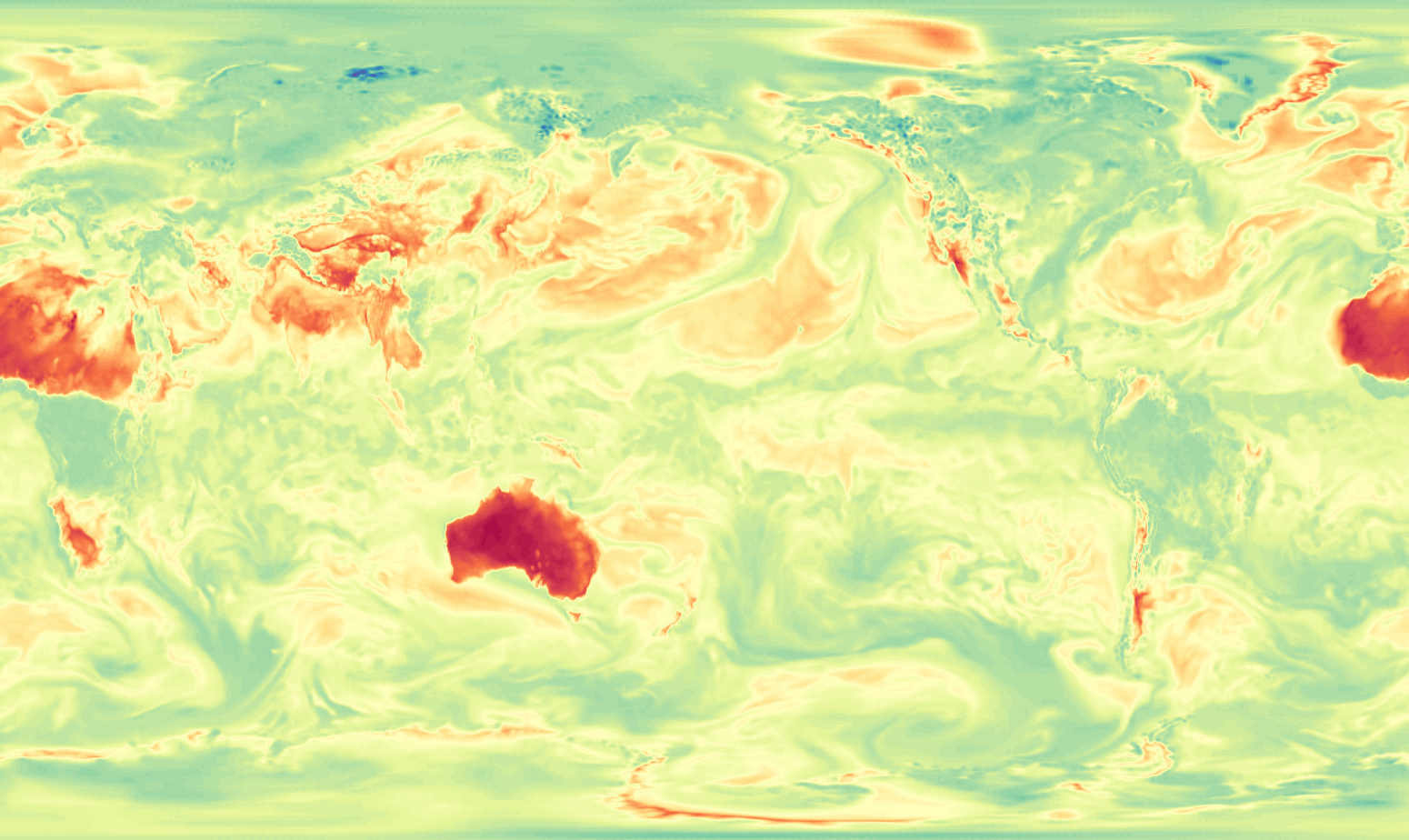} &
    \includegraphics[width=0.45\textwidth]{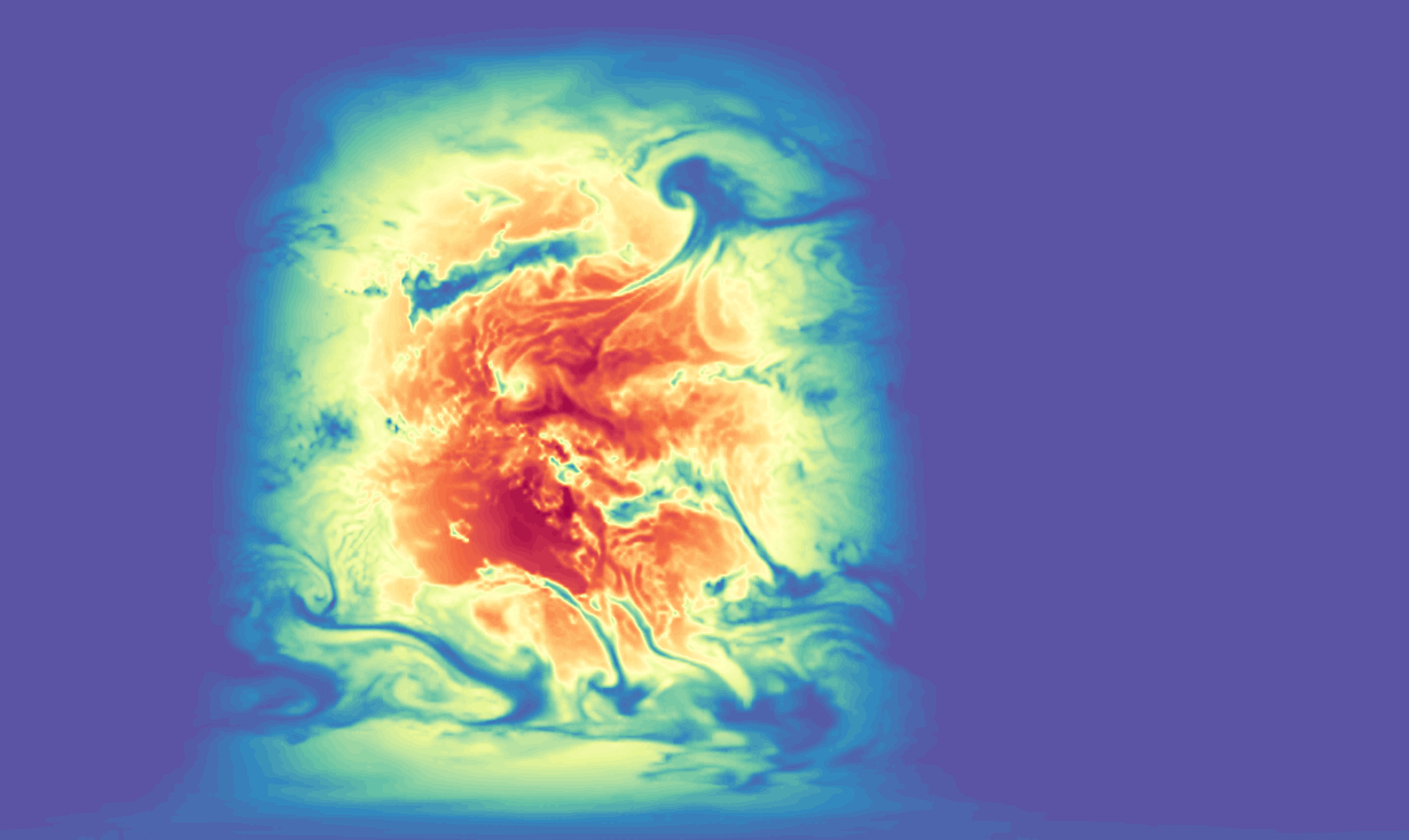} \\
    \small (e) 2m relative humidity prediction & \small (f) Solar radiation prediction \\
\end{tabular}
\caption{Various weather predictions by EPT-1.5, downsampled}
\label{fig:all_predictions}
\end{figure*}

The results for 10m and 100m wind speed as well as solar radiation and 2m temperature are collected in Tables \ref{tab:10m}, \ref{tab:2m}, \ref{tab:100w}, and \ref{tab:ssrd_skill}, 
while a visual overview of the current model performance is provided in Figures \ref{fig:10m_wind_speed}, \ref{fig:100m_wind_speed}, \ref{fig:2m_temperature}, and \ref{fig:6h_solar}. Finally, prediction samples of major weather variables of interest is depicted in Figure \ref{fig:all_predictions}.


Overall, the benchmarks reveal significant improvements and strong performance for \mbox{EPT-1.5} compared to its predecessor and current state-of-the-art models across multiple weather variables.

\subsection{Grid Performance}

\textbf{10m and 100m Wind Speed.}
\mbox{EPT-1.5} demonstrates strong performance in 10m wind speed prediction, marking a clear improvement over its predecessor, \mbox{EPT-1}. Notably, \mbox{EPT-1.5} outperforms HRES, GraphCast, FuXi and Pangu-Weather consistently across all forecast periods. While we could not compare \mbox{EPT-1.5} against other AI models for 100m wind speed prediction due their inability to predict this variable, \mbox{EPT-1.5} shows significant outperformance of \mbox{EPT-1} and IFS HRES. This achievement underscores \mbox{EPT-1.5}'s versatility and accuracy in wind speed prediction across various timescales.

\textbf{2m Surface Temperature.}
In surface temperature forecasting, \mbox{EPT-1.5} surpasses IFS HRES and \mbox{EPT-1} for most lead times, yet it lags behind GraphCast and FuXi. However, it is important to recognize that the initial versions of the product primarily focused on Wind and Solar Radiation.

\textbf{Solar Radiation.}
\mbox{EPT-1.5} shows a significant improvement over \mbox{EPT-1} in predicting 6-hourly surface (downward) solar radiation and significantly outperforms IFS HRES on all lead times. This capability is particularly noteworthy as it's a unique feature among current AI weather models, with neither GraphCast, FuXi, nor Pangu-Weather offering predictions for this variable. 

\subsection{Point Observation Results (Operational Model)}

\textbf{10m Wind Speed.}
The operational EPT-1.5\textsubscript{Op} demonstrates exceptional performance in predicting 10m wind speed, consistently outperforming IFS HRES across all forecast time ranges. Notably, both EPT-1.5\textsubscript{Op} and IFS HRES were initialized with the exact same data, emphasizing EPT-1.5\textsubscript{Op}’s superior ability to model and predict wind speed under identical initial conditions. This result underscores the advantage for daily operational usage.

\begin{figure}[t]
  \centering
  \includegraphics[width=\columnwidth]{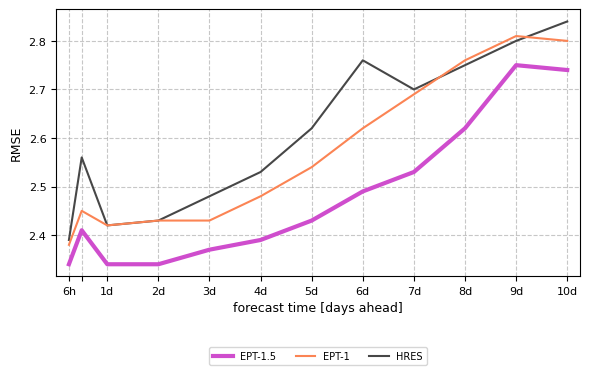}
  \caption{Comparison of 10m wind speed RMSE of the operational EPT-1.5\textsubscript{Op} (pink) vs. EPT-1 (orange) and HRES (black) at weather stations over Europe}
  \label{fig:point_10mwind}
\end{figure}

\begin{figure}[t]
  \centering
  \includegraphics[width=\columnwidth]{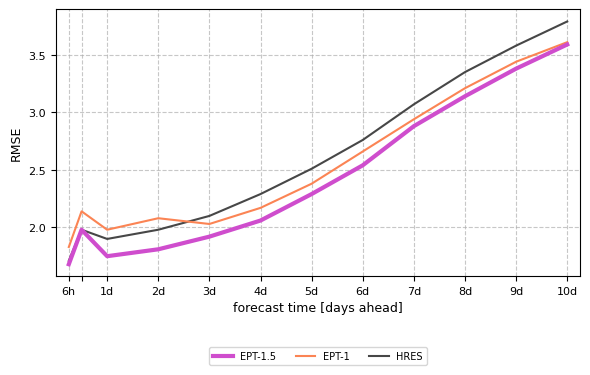}
  \caption{Comparison of 2m temperature RMSE of the operational EPT-1.5\textsubscript{Op} (pink) vs. EPT-1 (orange) and HRES (black) at weather stations over Europe}
  \label{fig:point_2m_temp}
\end{figure}

\textbf{2m Surface Temperature.}
EPT-1.5\textsubscript{Op} demonstrates remarkable accuracy in predicting 2m surface temperatures, consistently exceeding the performance of IFS HRES across all forecast ranges. Its precision is evident in both short-term and long-term forecasts, emphasizing EPT-1.5\textsubscript{Op}’s ability to handle varying environmental conditions with reliability. 

\begin{table*}[htbp]
\centering
\scriptsize 
\caption{Skill score for 10m wind speed by lead time for EPT variants and baseline models over Europe}
\begin{tabular}{c  c  c  c  c  c  c  c }
    \toprule
    Lead Time & {EPT-1.5\textsubscript{$\alpha$}} & {EPT-1.5\textsubscript{Base}} & {EPT-1} & {GraphCast} & {FuXi} & {Pangu-Weather} \\
    \midrule
    \midrule
    6 hours   & {20.8} & \textbf{21.7} & 9.1 & 17.9 & 12.4 & 0.62 \\
    12 hours  & \textbf{13.0} & 12.9 & 7.8  & 12.2 & 10.3 & 1.6 \\
    1 day     & \textbf{17.0} & 16.4 & 8.6 & 14.8 & 15.3 & 2.6 \\
    2 days    & \textbf{17.5} & 15.6 & 10.9  & 13.1 & 14.2 & 5.7 \\
    3 days    & \textbf{14.0} & 10.6 & 8.0  & 11.1 & 12.5 & 6.3 \\
    4 days    & \textbf{14.2} & 11.2 & 8.6  & 10.3 & 11.3 & 6.6 \\
    5 days    & \textbf{13.9}  & 11.0  & 8.7  & 9.6  & 9.9  & 5.3 \\
    6 days    & \textbf{13.5} & 9.0 & 8.6  & 9.7  & 1.7  & 6.1 \\
    7 days    & \textbf{9.9} & 6.5 & 5.5  & 8.0  & -0.13 & 5.6 \\
    8 days    & \textbf{8.4}  & 6.0 & 7.6  & 7.1  & -0.74 & 4.8 \\
    9 days    & 9.5  & 7.9 & \textbf{10.5}  & 6.1  & -2.1 & 3.4 \\
    10 days   & \textbf{7.8} & 1.6 & 7.7  & 6.5  & -2.8 & 3.7 \\
    \bottomrule
\end{tabular}
\label{tab:10m}
\end{table*}

\begin{table*}[htbp]
\centering
\scriptsize 
\caption{Skill score for 2m temperature by lead time for EPT variants and baseline models over Europe}
\begin{tabular}{c  c  c  c  c  c  c }
    \toprule
    Lead Time & {EPT-1.5\textsubscript{$\alpha$}} & {EPT-1} & {GraphCast} & {FuXi} & {Pangu-Weather} \\
    \midrule
    \midrule
    6 hours   & -47.0 & -62.8 & \textbf{-34.2} & -57.0 & -65.9 \\
    12 hours  & -8.2  & -19.0 & \textbf{-2.6}  & -30.0 & -32.6 \\
    1 day     & -0.9  & -10.0  & \textbf{6.5}   & -1.2  & -8.9  \\
    2 days    & 3.8   & -3.8  & \textbf{9.2}   & 2.2   & -4.3  \\
    3 days    & 7.3   & 0.2   & \textbf{8.3}   & 3.0   & -3.5  \\
    4 days    & 7.5   & 1.2  & \textbf{7.6}   & 4.1   & -1.9  \\
    5 days    & \textbf{6.6}  & 2.7  & 6.0   & 3.8   & -1.7  \\
    6 days    & \textbf{10.0}   & 5.3  & 6.0   & 10.2  & -0.7  \\
    7 days    & 6.9   & 3.2  & 6.4   & \textbf{12.8}  & 1.9   \\
    8 days    & 3.2   & -0.1  & 5.8   & \textbf{14.6}  & 2.0   \\
    9 days    & 4.5   & 3.0   & 4.6   & \textbf{16.6}  & 1.8   \\
    10 days   & 3.0  & 2.0   & 2.7   & \textbf{17.8}  & 0.1   \\
    \bottomrule
\end{tabular}
\label{tab:2m}
\end{table*}

\begin{table}[htbp]
\centering
\caption{Skill score for 100m wind speed by lead time for EPT-1.5\textsubscript{$\alpha$} and EPT-1 over Europe}
\begin{tabular}{c c c}
    \toprule
    Lead Time & EPT-1.5\textsubscript{$\alpha$} & EPT-1 \\
    \midrule
    6 hours   & \textbf{25.25} & 16.19 \\
    1 day     & \textbf{18.51} & 11.91 \\
    3 days    & \textbf{14.80} & 9.57  \\
    5 days    & \textbf{14.51} & 1.30  \\
    10 days   & 7.16  & \textbf{9.50}  \\
    \bottomrule
\end{tabular}
\label{tab:100w}
\end{table}

\begin{table}[htbp]
\centering
\caption{Short term skill score for solar radiation by lead time for EPT-1.5 and EPT-1 over Europe}
\begin{tabular}{c c c}
    \toprule
    Lead Time & EPT-1.5\textsubscript{Diagnostic} & EPT-1\\
    \midrule
    6 hours  & \textbf{20.44} & 13.19 \\
    1 day    & \textbf{10.53} & 2.14 \\
    3 days   & \textbf{6.25}  & 3.93 \\
    5 days   & 1.20  & \textbf{1.24} \\
    10 days  & 1.91  & \textbf{4.86} \\
    \bottomrule
\end{tabular}
\label{tab:ssrd_skill}
\end{table}

\begin{table}[htbp]
\centering
\caption{Comparison of the operational EPT-1.5\textsubscript{Op} vs. EPT-1 and HRES on 2m temperature on weather stations over Europe (RMSE)}
\begin{tabular}{c c c c}
    \toprule
    Lead Time & EPT-1.5\textsubscript{Op} & EPT-1 & IFS-HRES \\
    \midrule
    6 hours  & \textbf{1.67}  & 1.83 & 1.70  \\
    1 day & \textbf{1.74}  & 1.98 & 1.89  \\
    3 days   & \textbf{1.92}  & 2.03 & 2.11  \\
    5 days   & \textbf{2.29}  & 2.38 & 2.51  \\
    10 days  & \textbf{3.58}  & 3.61 & 3.79  \\
    \bottomrule
\end{tabular}
\label{tab:temp_point}
\end{table}

\begin{table}[htbp]
\centering
\caption{Comparison of the operational EPT-1.5\textsubscript{Op} vs. EPT-1 and HRES on 10m wind speed on weather stations over Europe (RMSE)}
\begin{tabular}{c c c c}
    \toprule
    Lead Time & EPT-1.5\textsubscript{Op} & EPT-1 & IFS-HRES \\
    \midrule
    6 hours  & \textbf{2.33}  & 2.38 & 2.38  \\
    1 day    & \textbf{2.34}  & 2.42 & 2.41  \\
    3 days   & \textbf{2.37}  & 2.43 & 2.48  \\
    5 days   & \textbf{2.48}  & 2.54 & 2.61  \\
    10 days  & \textbf{2.74}  & 2.80 & 2.83  \\
    \bottomrule
\end{tabular}
\label{tab:wind_point}
\end{table}
\section{Discussion}
\label{section:disc}

The EPT-1.5 model demonstrates significant advancements in  weather forecasting compared to its predecessor EPT-1 and other AI weather models. It provides state-of-the-art wind speed predictions and introduces solar radiation forecasting, addressing key needs in sectors such as energy production, gas, oil, and renewable energy. The model offers hourly global forecasts up to 20 days in advance and high-resolution ensemble predictions, delivering valuable information for energy production planning and grid management. 

In operational testing, EPT-1.5 consistently performs better than ECMWF HRES, the leading numerical weather model, across a range of forecast variables. These enhancements are particularly relevant for applications requiring accurate predictions of temperature, solar radiation, and wind patterns, which are crucial for the energy industry. The model's ability to generate rapid, high-accuracy forecasts indicates its potential for integration into real-time systems, supporting decision-making processes that rely on precise weather information.
\section{Outlook}
While \mbox{EPT-1.5} introduced a minor technical change over \mbox{EPT-1}, it delivered a significant boost in performance. The next major release of the EPT family, however, is set to bring a substantial leap in both technical innovation and performance, with even greater improvements in capabilities and generalization.

\section{Acknowledgements}
\label{section:ack}
We acknowledge the ECMWF for high-quality data like ERA5 and CERRA, and essential libraries like ecCodes. We thank the Max Planck Institute for Meteorology for their contributions to meteorological software tools that support our model development. We appreciate Stephan Rasp and the WeatherBench team for establishing a benchmark for evaluating AI weather models. We are grateful to the developers of the Zarr file format for enabling efficient storage and access to large datasets, and to Matthew Rocklin and the Dask development team for the Dask library, which has been instrumental in scaling our computational workflows. Finally, we thank our data partners and the broader meteorological community for their ongoing support in helping us evaluate and improve our models.
\bibliographystyle{IEEEtran}
\bibliography{IEEEabrv,refs}

\clearpage
\onecolumn
\section{Supplementary Material}
\label{section:appendix}

\vspace{1cm} 

\end{document}